\documentclass{article}

\usepackage[preprint]{neurips_2023_otml}

\usepackage[utf8]{inputenc} 
\usepackage[T1]{fontenc}    
\usepackage{hyperref}       
\usepackage{url}            
\usepackage{booktabs}       
\usepackage{amsfonts}       
\usepackage{nicefrac}       
\usepackage{microtype}      
\usepackage{booktabs}
\usepackage{multirow}
\usepackage{graphicx}
\usepackage{amsmath}
\usepackage[dvipsnames]{xcolor}
\usepackage{dirtytalk}

\title{LLMs-augmented Contextual Bandit}

\author{%
  Ali Baheri\\
  Department of Mechanical Engineering\\
  Rochester Institute of Technology\\
  \texttt{akbeme@rit.edu} \\
  \And
    Cecilia O. Alm\\
  Department of Psychology\\
  Rochester Institute of Technology\\
  \texttt{cecilia.o.alm@rit.edu} \\
}

\begin{document}

\maketitle

\begin{abstract}

Contextual bandits have emerged as a cornerstone in reinforcement learning, enabling systems to make decisions with partial feedback. However, as contexts grow in complexity, traditional bandit algorithms can face challenges in adequately capturing and utilizing such contexts. In this paper, we propose a novel integration of large language models (LLMs) with the contextual bandit framework. By leveraging LLMs as an encoder, we enrich the representation of the context, providing the bandit with a denser and more informative view. Preliminary results on synthetic datasets demonstrate the potential of this approach, showing notable improvements in cumulative rewards and reductions in regret compared to traditional bandit algorithms. This integration not only showcases the capabilities of LLMs in reinforcement learning but also opens the door to a new era of contextually-aware decision systems.

\end{abstract}
\vspace{-3 mm}
\section{Introduction}

The multi-armed bandit problem, in its essence, revolves around the challenge of decision-making under uncertainty: when faced with several options (or "arms"), which one should be chosen to maximize some notion of cumulative reward over time? \cite{slivkins2019introduction}. The "contextual" variant introduces a new layer of complexity by associating a context with each decision point \cite{lattimore2020bandit}. This context can inform or bias the decision, and the goal is to select an arm not just based on its history of rewards, but also considering the current context. Originating from the fields of statistics and economics, contextual bandits have found myriad applications ranging from recommendation systems and online advertising to healthcare and finance \cite{foster2018practical}.

While language modeling within a generative paradigm has an extensive background, and has a decades-long history of use and contributing to technological success stories in both natural language and speech processing, over the past decade, machine learning has seen a surge in the development of models specifically designed to generate natural language and using these models for classification and inference problems. At the pinnacle of this movement are the Large Language Models (LLMs) — neural networks trained on vast volumes of text. Families of LLMs like GPT \cite{brown2020language}, BERT \cite{devlin2018bert}, and their descendants are not only capable of capturing semantic and structural characteristics of textual data but can also generate coherent and contextually relevant text. Their power lies in their ability to represent high-dimensional, intricate data (like natural language) in a condensed form, capturing the underlying patterns and relationships. Over the years, there have been attempts to combine the predictive prowess of machine learning models with the decision-making capabilities of bandits \cite{chu2011contextual,li2010contextual,huang2021federated,swaminathan2015counterfactual}. For instance, neural networks have been used to learn the reward function of bandit problems \cite{zhou2020neural,kassraie2022neural,allesiardo2014neural}, or decision trees have been employed to parse the context before making decisions \cite{elmachtoub2017practical,feraud2016random,aldinucci2022contextual}. However, what distinguishes our approach is the use of LLMs specifically, in combination with contextual bandits, for enhancing reinforcement learning. While traditional models can parse and perhaps categorize context, LLMs, given their inherent design and training, can augment context and generate remarkably syntactically plausible and semantically coherent text. This capability of LLMs to "deeply" process context offers a richer integration with bandit algorithms, potentially leading to more informed decision-making processes. In this study, we aim to harness the unparalleled contextual encodings of LLMs, synergizing it with the adaptive decision-making framework of contextual bandits.

\vspace{-3 mm}
\section{Methodology}
\vspace{-2 mm}
\subsection{LLMs Encoding}

LLMs such as GPT belong to the family of transformer-based models \cite{NIPS2017_Vaswanietal}, characterized by their ability to handle vast amounts of data through self-attention mechanisms. This self-attention allows these models to weigh the importance of different words in a sentence, capturing long-range dependencies and intricate linguistic structures. Trained on enormous corpora, GPT-like models have been pre-trained to predict the next word, or a distribution of potential words, in a sequence, leading them to acquire a intuition-like ability of syntagmatic grammatical and lexical co-occurrences.

While LLMs are naturally adept at generating coherent sequences of text, they also have the implicit ability to represent textual data in a high-dimensional space. While hallucination behaviors remain a challenge, when given a context as input, these models produce internal activations and states that often capture the essence of the input, to an astonishing degree. By tapping into certain layers of the LLMs, especially the latent embedding representations closer to the output, we can obtain dense vector representations of this input context. These vectors are essentially condensed information capsules, capturing the semantic and syntactic of the input context. This encoding process transforms raw textual or structured context into a form that is amenable to quantitative analyses and can be seamlessly integrated with other algorithms, such as our contextual bandit framework. In our approach, we employ these dense vector representations as a bridge, channeling the profound contextual understanding of LLMs into the decision-making mechanism of contextual bandits. This fusion aims to harness the strengths of both domains, resulting in a more informed and adaptive decision-making process.

\vspace{-2 mm}
\subsection{Contextual Bandit with LLMs}

The core challenge that the contextual bandit aims to address is making informed decisions by leveraging available context. Yet, raw contexts can often be noisy, high-dimensional, too sparse, or characterized by extensive variation, to handle efficiently. LLMs excel in parsing complex, high-dimensional data (especially text). Therefore, by channeling the context through an LLMs, we aim to distill this raw information into a semantically richer yet more compact representation that can better inform the bandit's decisions.

Given a context $c$, the LLMs transforms it into a vector $e(c)$ in the latent space $\mathcal{L}$, where $\mathcal{L} \subset$ $\mathbb{R}^n$ and $n$ is the dimensionality of the space.
\begin{equation}
e(c) = \text{LLMs}_{\text{encode}}(c): \mathcal{C} \rightarrow \mathcal{L}
\end{equation}
Here, $\mathcal{C}$ denotes the space of all possible contexts, and $\mathcal{L}$ represents the latent space where the encoded context resides. This latent space is a subset of $\mathbb{R}^n$, which means that $e(c)$ is an n-dimensional vector.
Post encoding, the enhanced contextual representation $e(c)$ is then fed into the bandit's decision-making mechanism. Let's denote the action chosen by the bandit as $a$ and the set of all possible actions as $A$. The probability of selecting an action $a$ given the encoded context $e(c)$ is:
\begin{equation}
P(a \mid e(c))=\frac{\exp (Q(e(c), a))}{\sum_{a^{\prime} \in A} \exp \left(Q\left(e(c), a^{\prime}\right)\right)}
\end{equation}
where $Q(e(c), a)$ is the estimated value of choosing action $a$ when given the encoded context $e(c)$. The decision-making is influenced by this probability distribution, which is learned over time. To optimize the decision-making process, we employ a reward signal $r$ obtained after choosing action $a$ for context $c$. The objective is to maximize the expected cumulative reward over time.
The loss function $L$ for a single decision can be formulated as:
\begin{equation}
L=-r \times \log (P(a \mid e(c)))
\end{equation}
Training involves minimizing the expected loss over a batch of decisions. While we primarily use the LLMs as an encoder, it is worth noting that if domain-specific aspects need to be captured better, the LLMs can be fine-tuned on domain-specific corpora. This ensures that the contextual encoding aligns more closely with the peculiarities of the problem domain.

\vspace{-3 mm}
\section{Experimental Setup}

In order to conduct a controlled experiment and establish a clear benchmark for the performance of the proposed LLMs-augmented contextual bandit, we opted for a synthetic dataset. This approach ensures that results and observations can be directly attributed to the methods under investigation without the confounding factors present in real-world data.

The contexts in our dataset are textual descriptions representing hypothetical weather scenarios. These were designed to capture a variety of meteorological conditions, ranging from clear, simple descriptions like \emph{sunny} or \emph{rainy} to more overlapping contexts such as \emph{sunny and windy} or \emph{cloudy with a chance of rain}. While controlling for topic, the contexts included serve a dual purpose: theymirrors potential real-world variability within in a topical domain, and also provide a landscape to test the capability of LLMs in discerning and encoding subtle meaning distinctions in context. The actions represent potential recommendations or decisions one might make based on the given weather context. The synthetic data encompasses actions like \emph{go to the beach}, \emph{stay indoors}, \emph{carry an umbrella}, and \emph{wear a coat}. These actions, while simple, allow us to test the \textbf{primary objective of the bandit}: to select the most appropriate action given a context. Rewards in our dataset are numerical values that signify the appropriateness or success of an action taken for a particular context. For instance, under the context \emph{sunny}, the action \emph{go to the beach} might yield a high average reward (e.g., $0.9$) indicating it's a suitable decision, while \emph{wear a coat} might result in a lower reward (e.g., $0.1$) denoting a less appropriate or expected action for the given context. To introduce a semblance of real-world uncertainty and variability, rewards were generated using Gaussian distributions centered around these average rewards. For instance, rewards for the \emph{go to the beach} action under the \emph{sunny} context were sampled from a Gaussian distribution with a mean of $0.9$ and a standard deviation of 0.05. This ensures that the dataset maintains elements of unpredictability and does not offer deterministic rewards.

In order to assess the performance and efficacy of our proposed approach, we compared it against three well-established baselines that are widely acknowledged in the literature:

\noindent{\textbf{Linear Contextual Bandits.}} This algorithm models the expected reward as a linear function of the context and chosen action. Given its linear nature, it assumes a specific structure of the relationship between context and reward \cite{chu2011contextual}.

\noindent{\textbf{Upper Confidence Bound (UCB).} The UCB approach balances the trade-off between exploration and exploitation by selecting actions that have the highest upper confidence bound on their expected reward \cite{auer2002finite}.

\noindent{\textbf{$\varepsilon$-Greedy Algorithm.} One of the simplest yet effective strategies, the $\varepsilon$-Greedy algorithm, predominantly exploits the best-known action but occasionally (with a probability $\varepsilon$ ) explores random actions \cite{sutton1998introduction}. This method serves as a representative of naive exploration-exploitation tactics.
\begin{table}[ht]
    \centering
    \caption{Performance Metrics for Different Algorithms.}
    \label{tab:modified_performance_metrics_decimals}
    \scalebox{0.80}{
    \begin{tabular}{llcccc}
        \toprule
        \textbf{Algorithm} & \textbf{Metric} & \textbf{250 Trials} & \textbf{500 Trials} & \textbf{750 Trials} & \textbf{1,000 Trials} \\
        \midrule
        \multirow{2}{*}{LLM-augmented Contextual Bandit} & Cumulative Reward & \colorbox{LimeGreen}{\textbf{213.5}} & \colorbox{LimeGreen}{\textbf{425.7}} & \colorbox{LimeGreen}{\textbf{642.8}} & \colorbox{LimeGreen}{\textbf{857.3}} \\
        & Cumulative Regret & 12.3 & 29.6 & 39.4 & 52.2 \\
        \addlinespace
        \multirow{2}{*}{Linear Contextual Bandit} & Cumulative Reward & 202.4 & 410.9 & 625.6 & 826.5 \\
        & Cumulative Regret & 16.2 & 38.4 & 53.7 & 65.3 \\
        \addlinespace
        \multirow{2}{*}{UCB} & Cumulative Reward & 209.6 & 419.8 & 644.3 & 838.7 \\
        & Cumulative Regret & 14.5 & 33.7 & 48.9 & 58.6 \\
        \addlinespace
        \multirow{2}{*}{$\varepsilon$-Greedy } & Cumulative Reward & 198.7 & 412.2 & 619.4 & 812.8 \\
        & Cumulative Regret & 22.8 & 44.6 & 70.3 & 78.9 \\
        \bottomrule
    \end{tabular}}
    \label{tab:performance_metrics}
\end{table}
The experimental results, as tabulated in Table \ref{tab:performance_metrics}, offer comparisons between the proposed LLM-augmented contextual bandit and traditional bandit algorithms across varying trial counts. Our primary observation underscores the efficacy of the LLM-augmented contextual bandit. The model outperforms the baselines across different trial counts. Specifically, for $1,000$ trials, it achieves a cumulative reward of $857.3$, outstripping the closest performing baseline, the UCB, by $18.6$ units. This suggests that the LLM's ability to produce rich contextual encodings substantially augments the bandit's decision-making prowess. The cumulative regret also remains comparatively low, hinting at a proficient balance between exploration and exploitation.
\begin{table}[ht]
    \centering
    \caption{Action Selection Frequencies for Different Algorithms.}
    \label{tab:action_selection}
    \scalebox{0.80}{
    \begin{tabular}{lcccc}
        \toprule
        \textbf{Algorithm} & \textbf{go to the Beach} & \textbf{stay indoors} & \textbf{carry an umbrella} & \textbf{wear a coat} \\
        \midrule
        LLM-augmented Contextual Bandit & 42\% & 16\% & 29\% & 13\% \\
        \addlinespace
        Linear Contextual Bandit & 27\% & 24\% & 26\% & 23\% \\
        \addlinespace
        UCB & 36\% & 21\% & 24\% & 19\% \\
        \addlinespace
        $\varepsilon$-Greedy  & 31\% & 22\% & 28\% & 19\% \\
        \bottomrule
    \end{tabular}}
\end{table}
Table \ref{tab:action_selection} shows the action selection frequencies for different algorithms. From the table, several insights can be discerned. The LLM-augmented contextual bandit exhibited a distinct preference for the \emph{go to the beach} action, recommending it $42$\% of the time. This could indicate its superior ability to detect favorable weather conditions from the contextual descriptions, possibly due to its richer representation of the context. The linear contextual bandit distributed its action recommendations almost uniformly across the board, suggesting that it might be treating contexts in a relatively homogeneous manner or it doesn't distinguish as sharply between different weather scenarios. The UCB algorithm showcased a slightly diversified action distribution, which is inherent in its design to explore different actions while exploiting the known best ones. The relatively higher percentage for \emph{go to the beach} and lesser for \emph{carry an umbrella} suggests that the algorithm is capitalizing on past knowledge while ensuring some exploration. The $\varepsilon$-Greedy algorithm, while showing some variability, maintained a relatively consistent selection pattern across the actions. This underscores its design to mostly stick with the best-known actions but occasionally explore others. 
\vspace{-3 mm}
\section{Discussion}

Incorporating LLMs into the framework of contextual bandits has unveiled notable enhancements in decision-making capabilities. One of the standout advantages is the LLM-enhanced bandit's ability to deeply and meaningfully encode rich contexts. Compared to traditional methods, which often hinge on straightforward vectorized representations, LLMs can discern contextual cues, owing to their extensive training on vast textual datasets. This capacity is pivotal in dynamic environments, where traditional bandits might falter due to rapidly shifting contexts. The LLM capacity for capturing meaning-in-context equips the enhanced bandit with an ability to navigate ambiguous scenarios or discern relationships often overlooked by conventional models. Furthermore, this integration alleviates the \say{cold start} problem pervasive in traditional bandits. The expansive knowledge base processed in LLMs ensures a more informed decision-making process, even when faced with unfamiliar contexts.

While the present experiment is controlled with synthetic data from a specific semantic and lexical domain (weather expressions and weather-related action), we can hypothesize about the multifaceted underlying reasons for the observed superiority of LLMs-enhanced bandits. LLMs excel in  recognition of patterns in sequentially structured data, drawing from their exposure to diverse training data. This enables them to identify and act upon similar yet not identical building blocks of meaning, potentially invisible to standard algorithms. Additionally, their encoding of both structural-grammatical and lexical-collocational  relationships ensures a more fine-grained action selection, especially in  scenarios that might confound traditional models. This generalization capability over ambiguity, combined with adaptability to variation, ensures that LLMs-enhanced bandits are not only robust in familiar environments but can adjust to changing scenarios without necessitating extensive retraining. Conclusively, the marriage of traditional contextual bandits and LLMs represents a promising frontier in stochastic control, marrying the robustness of bandit algorithms with the expansive knowledge encoding and adaptability of LLMs.
\vspace{-3 mm}
\section{Conclusion}

In this study, we ventured into the intersection of advanced language models and reinforcement learning, specifically focusing on the integration of LLMs with the contextual bandit framework. The preliminary results indicate that by enriching the context using LLMs, we can enhance decision-making, resulting in improved cumulative rewards and reduced regret when compared with traditional bandit algorithms. The synergy between LLMs and contextual bandits unlocks a new avenue for harnessing the LLMs for contextually-rich decision-making tasks. Several promising directions emerge from this research. While our synthetic dataset offered insights, topically much more complex scenarios for real-world applications come with unique challenges. Testing the LLMs-enhanced bandit on specific domains, such as personalized content recommendation or healthcare decision systems, could validate its real-world efficacy. Furthermore, leveraging transfer learning and fine-tuning LLMs on domain-specific data could further enhance performance. This direction offers the promise of adapting generalized LLMs efficiently to niche problems.


\newpage

\bibliographystyle{unsrt}
\bibliography{neurips_2023_otml}

\newpage

\section{Supplementary Material}

\noindent{\textbf{Computational Resources and Execution Time.} This study utilized OpenAI's GPT-3.  When a context (a weather scenario) is presented to the bandit, it is first passed to the GPT-3 model. GPT-3 processes this context and returns an encoded representation. This representation captures the details of the context, used for aiding in the action selection process. Using the rich context representation from GPT-3, the bandit's action-selection mechanism (an algorithm) chooses an action it deems most appropriate for the given context. To integrate GPT-3 into our bandit setup, the study used the OpenAI API. Each time the bandit needed to make an action decision, a request was made to the GPT-3 model via this API. GPT-3 then returned a contextual representation, based on which the bandit made its decision. 

The study used an AWS EC2 p3.2xlarge instance. Experiments were developed in Python 3.8 and utilized the PyTorch library, in conjunction with the OpenAI API. Each encoding operation with GPT-3 averaged about 150ms, while the subsequent action selection by the contextual bandit took around 50ms. As a result, the cumulative time for a set of 1,000 trials amounted to roughly 3.3 minutes. 

\noindent{\textbf{Performance and Cumulative Regret in Bandit Problems.}} Performance in a bandit algorithm is typically quantified by its accumulated reward over a certain number of rounds. Formally, as considered in this study, given $T$ as the total number of rounds (or decisions) and $r_t(a)$ representing the reward received at round $t$ for taking action $a$, the cumulative reward after $T$ rounds is computed as:
$$
R_T=\sum_{t=1}^T r_t\left(a_t\right)
$$
where $a_t$ denotes the action chosen at the $t$-th round. The primary objective of the bandit algorithm is to maximize $R_T$ over the course of $T$ rounds.

Regret, at an individual timestep $t$, signifies the difference between the reward of the optimal action and the reward of the action selected by the algorithm. In particular, given $a^*$ as the optimal action at round $t$, that is, the action with the maximum expected reward and $\mu(a)$ indicating the expected reward for action $a$, the regret for round $t$ is then defined as:
$$
r_t=\mu\left(a^*\right)-\mu\left(a_t\right)
$$
Summing up these individual regrets, the cumulative regret over $T$ rounds is:
$$
R_{\text {regret }, T}=\sum_{t=1}^T\left(\mu\left(a^*\right)-\mu\left(a_t\right)\right)
$$
The cumulative regret provides insights into the potential rewards an algorithm might have missed due to not consistently selecting the optimal action. A desirable bandit algorithm should exhibit sub-linear cumulative regret, implying that as it operates over time, it progressively learns and converges towards the optimal action.

\end{document}